\RequirePackage{fix-cm}
\documentclass[smallextended,natbib]{svjour3}       % onecolumn (second format)
\usepackage{graphicx}
\usepackage{amssymb}
\usepackage{amsmath}
\usepackage{booktabs}
\newcommand{\ie}{\emph{i.e.}}
\newcommand{\eg}{\emph{e.g.}}
\newcommand{\wrt}{\emph{w.r.t}}

\journalname{International Journal of Computer Vision}
\begin{document}
\title{Deep Perceptual Mapping for Cross-Modal Face Recognition%\thanks{Grants or other notes
}


\author{M. Saquib Sarfraz         \and
        Rainer Stiefelhagen %etc.
}


\institute{M.S. Sarfraz \and R. Stiefelhagen \at Institute of Anthropomatics \& Robotics, Karlsruhe Institute of Technology (KIT). \\
Karlsruhe, Germany\\Tel.: +49-721-608-41694, Fax: +49-721-608-45939,  \email{saquib.sarfraz@kit.edu}           %  \\
%             \emph{Present address:} of F. Author    if needed
 }


\maketitle
\begin{abstract}
Cross modal face matching between the thermal and visible spectrum is a much desired capability for night-time surveillance and security applications. Due to a very large modality gap, thermal-to-visible face recognition is one of the most challenging face matching problem. In this paper, we present an approach to bridge this modality gap by a significant margin. Our approach captures the highly non-linear relationship between the two modalities by using a deep neural network. Our model attempts to learn a non-linear mapping from the visible to the thermal spectrum while preserving the identity information. We show substantive performance improvement on three difficult thermal-visible face datasets. The presented approach improves the state-of-the-art by more than 10\% on the UND-X1 dataset and by more than 15-30\% on the NVESD dataset in terms of Rank-1 identification. Our method bridges the drop in performance due to the modality gap by more than 40\%. 
%\keywords{Heterogeneous face recognition \and thermal face recognition \and night-time surveillance}
\end{abstract}

\section{Introduction}
\label{sec:intro}
Face recognition, mainly, has been focused in the visible spectrum. This pertains to a large number of applications from biometrics, access control systems, social media tagging to person retrieval in multimedia. Among the main challenges in visible face recognition, the different lighting/illumination condition has proven to be one of the big factors for appearance change and performance degradation. Many prior studies such as \cite{li2007, socolinsky2002, nicolo2012, Klare13} have stated better face recognition performance in the infra-red spectrum because it is invariant to ambient lighting. Relatively recently, few efforts have been devoted in the cross-modal face recognition scenarios, where the objective is to identify a person captured in infra-red spectrum based on its stored high resolution visible face image. The motivation for this lies in the night-time or low light surveillance tasks where the image is captured discretely or covertly through active or passive infra-red sensors. In fact there does not exist a reliable solution for matching thermal faces, acquired covertly in such conditions, against the stored visible database such as police mugshots. This poses a significant research gap for such applications in aiding the law enforcement agencies. In the infra-red spectrum, thermal signatures emitted by skin tissues can be acquired through passive thermal sensors without using any active light source. This makes it an ideal candidate for covert night-time surveillance tasks.

\begin{figure}
\centering
    \includegraphics[width=0.95\textwidth]{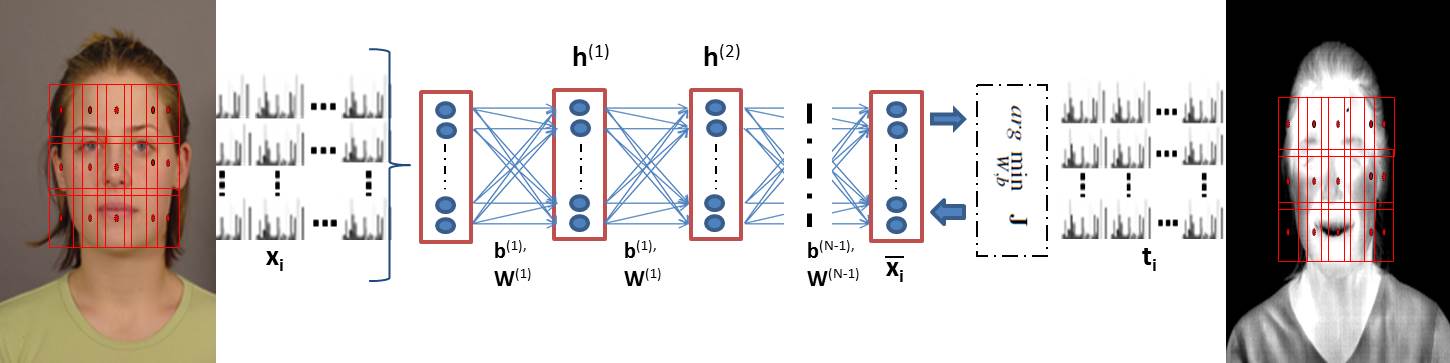}
    \caption{Deep Perceptual Mapping (DPM): densely computed features from the visible domain are mapped through the learned DPM network to the corresponding thermal domain.}
    \label{fig:1}
    \end{figure}
    
As opposed to the visible spectrum wavelength (0.35$\mu m$ to 0.74$\mu m$), the infra-red spectrum lies in four main ranges. Near infra-red `NIR' (0.74$\mu m$-1$\mu m$), short-wave infra-red `SWIR' (1-3$\mu m$), mid-wave infra-red `MWIR' (3-5$\mu m$) and long-wave infra-red `LWIR' (8-14$\mu m$). Since the NIR and SWIR bands are reflection dominant, they are more close to the visible spectrum. The MWIR and LWIR are the thermal spectrum and are emission dominant i.e. dependent on material emissivity and temperature. Skin tissue has high emissivity in both the MWIR and LWIR spectrum. Because of this natural difference between the reflective visible spectrum and sensed emissivity in the thermal spectrum, images taken in the two modalities are very different and have a large modality gap. This hinders reliable face matching across the two domains. It is, perhaps, for this reason that most of the earlier studies, aiming at cross-modal face recognition, rely only on visible-to-NIR face matching. While achieving very good results, NIR imaging use an active light source that makes it redundant for covert night-time surveillance. More recently, some attempts have been made in thermal-to-visible face recognition, indicating a significant performance gap due to the very challenging nature of the problem and the large modality gap.

In this paper, we seek to bridge this gap by trying to directly model the highly non-linear mapping between the two modalities. Our contribution is a useful model, based on a feed-forward deep neural network, and its effective design steps in order to map the perceptual differences between the two modalities while preserving the identity information. We show that this mapping can be learned from relatively little training data and that it works quite well in practice. Our model tries to learn a non-linear regression function between the visible and thermal data, where this data is comprised of densely pooled feature vectors from the images in the corresponding domain. Figure \ref{fig:1} summarises our approach. The learned projection matrices capture the non-linear relationship well and are able to bring the two closer to each other. Another contribution is to further the best published state-of-the-art performance on very challenging datasets i.e., the University of Notre Dame's `UND-X1' LWIR-visible dataset by more than 10\% and on Night Vision Electronics \& US Army Research Lab's `NVESD' MWIR-visible and LWIR-visible dataset by 18\% and 35\% respectively. Our results show that this accounts for bridging the performance gap due to the modality difference by more than 40\%. To the best of our knowledge, this is the first attempt in using deep neural networks to bridge the modality gap in cross-modal face recognition. Figure \ref{fig:1} provides an overview of the approach. 

This is the extended version of our previously published conference paper (\cite{sarfraz2015}) including a more thorough evaluation and analysis of the method. We will start by discussing some of the related work in section \ref{related}. Section \ref{DPM} will detail the presented approach. We will conclude in section \ref{Disc} after presenting detailed experiments, results and implementation details in section \ref{EXP}. 

\section{Related Work}
\label{related}
One of the very first comparative studies on visible and thermal face recognition was performed by \cite{socolinsky2002}. They concluded that "LWIR thermal imagery of human faces is not only a valid biometric,  but almost surely a superior one to comparable visible imagery." A good survey on single model and cross modal face recognition methods can be found in \cite{zhou2014recent}.

In the cross-modal (infra-red-visible) face recognition scenario, most of the earlier efforts focus only in the NIR to Visible matching. One of the first investigation by \cite{yi2007} uses Linear discriminant analysis (LDA) and canonical correspondence analysis to perform linear regression between NIR and visible images. A number of approaches build on using local feature descriptors to represent the face.   \cite{liao2009} first used this approach on NIR to visible face recognition by processing face images with a difference of Gaussian (DoG) filter, and encoding them using multiblock local binary patterns. Gentle AdaBoost feature selection was used in conjunction with LDA to improve the recognition accuracy. \cite{Klare2010} followed this work on NIR to visible face recognition by also incorporating SIFT feature descriptors and an LDA scheme. \cite{lei2009} applied coupled spectral regression for NIR to visible recognition. Few methods have also focused on SWIR-to-visible face recognition \cite{ross2010}, \cite{nicolo2012}. NIR or SWIR to visible face matching produces relatively better results as both the modalities are very similar because of the small spectral gap. Because of their limited use in the night-time surveillance applications, a much needed research focus is required in the thermal to visible matching domain.

Only recently some interest in the thermal to visible face matching has emerged. In the thermal domain, most of these methods are evaluated in the MWIR to visible scenario.  These methods employ similar techniques based on local features as in NIR to visible. A nice departure was proposed by \cite{li2008} which is the only known method to perform recognition by trying to synthesize an equivalent visible image from the thermal counterpart. They provided some visual results and evaluated the method in a small hand collected data indicating an unsatisfactory performance. Another interesting local feature based approach has been put forward by \cite{Klare13}, using local kernel prototypes. Their idea is to represent the two modalities by storing only similar vectors as prototypes during training using a similarity kernel space termed as prototype random subspace. They tested the method in different cross modal scenarios including MWIR to visible face recognition. \cite{bourlai2012} proposed a MWIR-to-visible face recognition system based on simple preprocessing and feature matching. More recently \cite{hu2014} used a game theoretic based partitioning of the images to extract LBP type features and used SVM kernel matching for MWIR to visible face recognition achieving good results. A similar performance on the same dataset has been obtained in a very recent proposal \cite{hu2015} using partial least square `PLS' based mapping between the MWIR and visible domain. MWIR sensors acquire higher resolution images than the LWIR. They, however, also operate at lower emissivity response to sense than the LWIR. The most difficult acquisition scenario is the usage of LWIR thermal images. Because of the long wave range sensing in LWIR, the images produced are usually of considerably lower resolution making it more challenging to match. On the other hand, pertaining to their higher emissivity sensing range they are also well suited to operate in total darkness. Attempts have been made to address face matching in the LWIR to visible domain. \cite{choi2012} presented a PLS based regression framework to model the large modality difference in the PLS latent space. The most recent and best state-of-the-art results on the same dataset have been achieved by \cite{hu2015}. They used a discriminant PLS based approach by specifically building discriminant PLS gallery models for each subject by using thermal cross examples. They achieved a rank-1 identification of 50.4\% using two gallery images per subject and testing against all the probe LWIR thermal images. \cite{Chen2015} put forward an interesting approach centred on learning common subspaces between visible and thermal facial images. The common subspaces are learnt via successive subspace learning process using factor analysis and common discriminant analysis on the image patches from both the domains. The projected vectors on these learnt subspaces are then directly matched. They provide convincing results on a private dataset. Finally, for facial feature detection in thermal images,\cite{mostafa2013} presented a good method based on Haar features and Adaboost. They also performed  face recognition experiments, but used both the visible and thermal modalities in the gallery and probe set.

Deep and large neural networks have exhibited impressive results for visible face recognition. Most notably, Facebook (\cite{taigman2014}) and Google (\cite{schroff2015facenet}) showed excellent results using convolutional neural network to learn the face representation. \cite{hu2014discriminative} employ a feed forward network for metric-learning using pairs of same and different visible images. In contrast, we use a  network to learn the non-linear regression projections between visible and thermal data without trying to learn any distance metric or image representation. A similar objective is employed in the domain adaptation based approaches e.g., \cite{ganin2014}, where the idea is to use the knowledge of one domain to aid the recognition in another image domain. Our model, while trying to learn a non-linear regression mapping, is simple and different from these in terms of the objective function and the way the network is used. We use a fully-connected feed forward network with the objective of directly learning a perceptual mapping between the two different image modalities. 

\section{Deep Perceptual Mapping (DPM)}
\label{DPM}
The large perceptual gap between thermal and visible images is both because of the modality and resolution difference. The relationship between the two modalities is highly non-linear and difficult to model. Previous approaches try to use non-linear function mapping e.g., using the kernel trick or directly estimating the manifold where the two could lie closer together. Since such a manifold learning depends on defining the projections based on a given function, they are highly data dependent and require a good approximation of the underlying distribution of the data. Such a requirement is hard to meet in practice, especially in the case of thermal-visible face recognition. Here, deep neural networks can learn, to some extent, the non-linear mapping by adjusting the projection coefficients in an iterative manner over the training set. As the projection coefficients are learned and not defined, it may be able to capture the data specific projections needed to bring the two modalities closer together. Based on this foresight we attempt to learn such projections by using a multilayer fully-connected feed forward neural network. 

\subsection{The DPM Model}
The goal of training the deep network is to learn the projections that can be used to bring the two modalities together. Typically, this would mean regressing the representation from one modality towards the other. Simple feed forward neural network provides a good fitting architecture for such an objective. We construct a deep network comprising $N+1$ layers with $m^{(k)}$ units in the $k$-th layer, where $k=1, 2, \cdots, N$. For an input of $x\in \mathbb{R}^{d}$, each layer will output a non-linear projection by using the learned projection matrix $\mathbf{W}$ and the non-linear activation function $g(\cdot)$. The output of the $k$-th hidden layer is $h^{(k)}=g(\mathbf{W}^{(k)}h^{(k-1)}+\mathbf{b}^{(k)})$, where $\mathbf{W}^{(k)}\in \mathbb{R}^{m^{(k)} \times m{(k-1)}}$ is the projection matrix to be learned in that layer, $\mathbf{b}^{(k)}\in \mathbb{R}^{m^{(k)}}$ is a bias vector and $g:\mathbb{R}^{m^{(k)}} \mapsto \mathbb{R}^{m^{(k)}}$ is the non-linear activation function. Note, $h^{(0)}$ is the input. Similarly, the output of the most top level hidden layer can be computed as:
\begin{equation}
\label{eq1}
\mathbf{H}(x)=h^{(N)}=g(\mathbf{W}^{(N)}h^{(N-1)}+\mathbf{b}^{(N)})
\end{equation}
where the mapping $\mathbf{H}:\mathbb{R}^{d} \mapsto \mathbb{R}^{m^{(N)}}$ is a parametric non-linear perceptual mapping function learned by the parameters $\mathbf{W}$ and $\mathbf{b}$ over all the network layers.
Since the objective of the model is to map one modality to the other, the final output layer of the network employs a linear mapping to the output in Equation \ref{eq1} to  obtain the final mapped output:
\begin{equation}
\label{eq2}
\bar{x}=s(\mathbf{W}^{(N+1)}\mathbf{H}(x))
\end{equation}
where $\mathbf{W}^{(N+1)} \in \mathbb{R}^{m^{(N)} \times d}$ and $s$ is the linear mapping. Equation \ref{eq2} provides the final perceptually mapped output $\bar{x}$ for an input $x$.
To determine the parameters $\mathbf{W}$ and $\mathbf{b}$ for such a mapping, our objective function must seek to minimize the perceptual difference between the visible and thermal training examples in the least mean square sense. We, therefore, formulate the DPM learning as the following optimization problem.
\begin{equation}
\label{eq3}
arg ~ \min_{W,b} ~~~ \mathbf{J}=\frac{1}{M}\sum_{i=1}^M (\bar{x}_i - t_i)^2 + \frac{\lambda}{N} \sum_{k=1}^N(\|\mathbf{W}^{(k)}\|_F^2 + \|\mathbf{b}^{(k)}\|_2^2) 
\end{equation}

The first term in the objective function corresponds to the simple squared loss between the network output $\bar{x}$ given the visible domain input and the corresponding training example $t$ from the thermal domain. The second term in the objective is the regularization term with $\lambda$ as the regularization parameter. $\|\mathbf{W}\|_F$ is the Frobenius norm of the projection matrix $\mathbf{W}$. Given a training set $\mathbf{X}=\left\{x_1,x_2,\cdots,x_M\right\}$ and $\mathbf{T}=\left\{t_1,t_2,\cdots,t_M\right\}$  from visible and thermal domains respectively, the objective of training is to minimize the function in equation \ref{eq3} with respect to the parameters $\mathbf{W}$ and $\mathbf{b}$.
\\[0.5\baselineskip]\noindent\textbf{The DPM Training:} There are some important design considerations for a meaningful training of the DPM network to provide an effective mapping from visual to thermal domain. First, the model is sensitive to the number of input and output units. A high dimensional input would require a very high amount of training data to be able to reasonably map between the two modalities. We propose to use the densely computed feature representations from overlapping small regions in the images. This proves very effective, as not only the model is able to capture the differing local region's perceptual differences well but also alleviate the need of large training images and nicely present the input in relatively small dimensions. The training set $\mathbf{X}$ and $\mathbf{T}$, then, comprises of these vectors coming from the corresponding patches from the images of the same identity. Note, only using the corresponding images of the same identity ensures that the model will only learn the differences present due to the modality as the other appearance parameters \eg identity, expression, lighting etc., would most likely be the same. The network is trained, over the training set, by minimizing the loss in Equation \ref{eq3} \wrt the parameters. Figure \ref{fig:1} encapsulates this process. By computing the gradient of the loss $J$ in each iteration, the parameters are updated by standard back projection of error using stochastic gradient descent `SGD' method of \cite{glorot2010}. For the non-linear activation function $g(z)$ in the hidden layers, we have used the hyperbolic tangent `$tanh$' as it worked better than other functions \eg sigmoid and ReLU on our problem. $tanh$ maps the input values between -1 to 1 using $g(z)= \frac{e^{2z}-1}{e^{2z}+1}$.

Before commencing training, the parameters are initialized according to a uniform random initialization. As suggested in \cite{glorot2010}, the bias $\mathbf{b}$ is initialized as $0$ and the $\mathbf{W}$ for each layer is initialized using the uniform distribution $\mathbf{W}^{(k)}\sim  U\left[-\frac{\sqrt6}{\sqrt{m^{(k)}+m^{(k-1)}}},\frac{\sqrt6}{\sqrt{m^{(k)}+m^{(k-1)}}}\right]$, where $m^{(k)}$ is the dimensionality (number of units) of that layer.   
\subsection{Thermal-Visible Face Matching}
After obtaining the mapping from visible to thermal domain, we can now pose the matching problem as that of comparing the thermal images with that of mapped visible data. Specifically, the mapped local descriptors from overlapping blocks of the visible gallery images are concatenated together to form a long vector. The resulting feature vector values are $L_2$-normalized and then matched with the similarly constructed vector directly from the probe thermal image. Note that, only visible gallery images are mapped through the learned projections using Equations \ref{eq1} and \ref{eq2} whereas the densely computed vectors from thermal images are directly concatenated into the final representation vector.
 
The presented set-up is ideal for the surveillance scenario as the gallery images can be processed and stored offline while at test time no transformation and overhead is necessary. As we will show in the next section, without any computational overhead the probes can be matched in real-time. Note, however, the  presented DPM is independent of the direction of mapping. We observed only slight performance variations using the opposite thermal to visible mapping.

The identity of the probe image is simply determined by computing the similarity with each of the stored gallery image vectors and assigning the identity for which the similarity is maximum. Since the vectors are $L_2$-normalized, the cosine similarity can simply be computed by the \textit{dot} product.   
\begin{equation}
d(\bar{x_i},t_j)=\bar{x_i}\cdot t_j \qquad\qquad \forall i=1,2,\cdots ,G
\end{equation}
where $t_j$ is the $j$-th constructed probe thermal image vector and $G$ is the total number of stored gallery vectors. Computationally, this is very efficient as at the test time, each incoming probe image can be compared with the whole gallery set $\mathbf{\bar{X}}$ by a single matrix-vector multiplication.

\section{Experimental Results}
\label{EXP}
\begin{figure}
    \includegraphics[width=\textwidth]{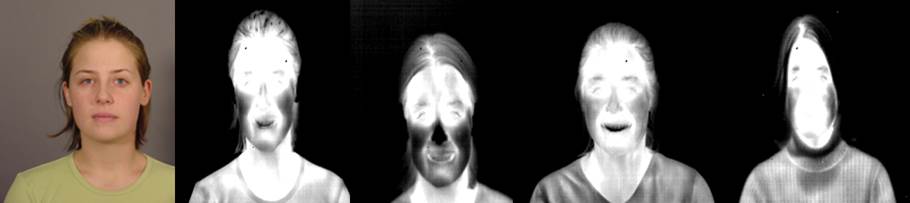}
    \caption{Problem: Matching the thermal to the stored high resolution visible image. A wide modality gap exists between the visible and the thermal images of the same subject. Thermal images depict typical image variations, present, due to lighting, expressions and time-lapse.}
    \label{fig:2}
\end{figure}

In this section we present the evaluation of the proposed approach on the difficult thermal (LWIR \& MWIR)-to-visible face recognition scenario. We report evaluations using typical identification and verification settings. The evaluation results, assessing the proposed DPM mapping, and the implementation details are discussed in the following subsections. 
\subsection{Database Description}
Only few publicly available datasets include thermal and corresponding visible facial acquisitions. Among these, we use three very challenging datasets to evaluate the performance of the proposed cross-modal face matching method: UND Collection X1, Carl Database, and NVESD dataset. Each database was collected with different sensors containing image acquisitions under different experimental conditions e.g., different lighting, facial expressions, time-lapse, physical exercise and subject-to-camera range.
\\[0.5\baselineskip]\noindent\textbf{UND Collection X1:} University of Notre Dame's UND collection X1 (\cite{chen2005ir}) is a challenging LWIR and visible facial corpus. The dataset was collected using a Merlin uncooled LWIR sensor and a high resolution visible color camera. The resolution of the visible images is $1600\times1200$ pixels and the acquired LWIR thermal images is $312\times239$ pixels. This depicts the challenge not only due to the large modality gap but also due to a large resolution difference. The dataset includes images in three experimental settings: expression (neutral, smiling, laughing), lighting change (FERET lighting and mugshot) and time-lapse (subjects acquired in multiple sessions over time). Following the protocol used by the recently published competitive proposals on this dataset \cite{hu2015, choi2012}, we use all the available images in these three settings of the thermal probes to match against a single or multiple available high resolution visible gallery image/s. The dataset contains $4584$ images of 82 subjects distributed evenly in visible and thermal domain. To compare our results, we use the exact same training and testing partitioning as used in the previous methods. The pool of subjects is divided into two equal subsets: visible and thermal images from 41 subjects are used for the gallery and probe images (partition A), while the remaining 41 subjects were used to train the DPM model. Note that the training and test sets are disjoint both in terms of images and subject's identities. All thermal images from partition-A are used as probes to compute recognition performance. The images are aligned based on provided eye and mouth locations and facial regions are cropped to $110\times150$ pixels. Figure \ref{fig:2} shows some sample visible and corresponding thermal images of one subject in the database.
\\[0.5\baselineskip]\noindent\textbf{Carl Database:}
The Carl dataset (\cite{carlDataset}) is a relatively recently collected dataset containing images of 41 subjects acquired in visible, near-infrared and thermal (LWIR) spectrum. The dataset collects images under three illumination conditions (natural, artificial and infrared lighting) in 4 different sessions over time. The images primarily contain challenging appearance variations due to different illuminations, expressions, time-lapse and low image resolution. We use the thermal and visible corpus of the data which contains images across different lighting and acquisitions sessions totalling to 41x5x3x4x2=4920 images, for 41 subjects, 3 illumination conditions (5 image/subject/illumination), 4-sessions in both visible and thermal domain. The resolution of the visible images is $640\times480$ and thermal (LWIR) images is $160\times120$ pixels. The segmented face images (output of a face detector) are provided by the authors. The face images, resized to $100\times145$ pixels, are not aligned w.r.t to eyes or nose position and therefore adds to the recognition challenge pertaining to misalignment errors. Since no earlier studies evaluate the full scale dataset in thermal to visible scenario, we provide the results to benchmark the full dataset in the following setting. We split the dataset equally with respect to the subjects. We use images of 20 subjects as the training set (1200 visible and 1200 thermal) and the rest as the probe. The training and test sets are, therefore, disjoint in terms of both subject's identities and images. Note, the probe set contains all the available thermal images of 21 subjects (from 3 illumination conditions in 4 sessions) totalling to 1260 images. The gallery set comprises visible images (1/subject and 2/subject from session 1, natural illumination) of all the 41 subjects. This protocol will ensure an effective evaluation on the challenging thermal image variations due to illumination, time lapse and expressions against a large gallery.
\\[0.5\baselineskip]\noindent\textbf{NVESD Dataset:}
The NVESD dataset (\cite{NVESD}) was acquired as a joint effort between the Night Vision Electronic Sensors Directorate of the U.S. Army Communications-Electronics Research and the U.S. Army Research Laboratory (ARL). We use the thermal subset of the data containing images in both MWIR and LWIR spectrum. The NVESD dataset primarily examined two conditions in both spectrum: physical exercise (fast walk) and subject-to-camera range (1 m, 2 m, and 4 m). The dataset contains images of a total of 50 subjects, where images of 25 subjects are acquired in both before exercise and after exercise condition across all three ranges. The image resolution for the thermal sensors (MWIR and LWIR) is considerably higher at $640\times480$ pixels. The total number of thermal images in NVESD is 900 (450 MWIR, 450 LWIR) and the visible images are 450. The training set,  correspond to subjects whose images are only acquired in before exercise condition,  comprises 300 thermal images (150 MWIR, 150 LWIR) and 150 visible images. We train two separate models for MWIR-visible and LWIR-visible mapping using these training images. 

Following the protocol in recently published state-of-the-art results on this dataset by \cite{hu2015}, we use the 25 subject subset (before and after exercise across all three ranges) as the probe/test set to examine the performance. The total number of probe thermal images is 600 (300 MWIR \& 300 LWIR).
The gallery set comprises the visible images of all the 50 subjects from before exercise condition at 1 m range. This protocol helps evaluate the performance in the presence of changed thermal signature (due to exercise) and subject-to-camera distance against the standard mugshot visible gallery image/s. The face images are normalised and aligned to $272\times322$ pixels as in \cite{hu2015} in order to have a fair comparison.

\begin{figure}
    \includegraphics[width=\textwidth]{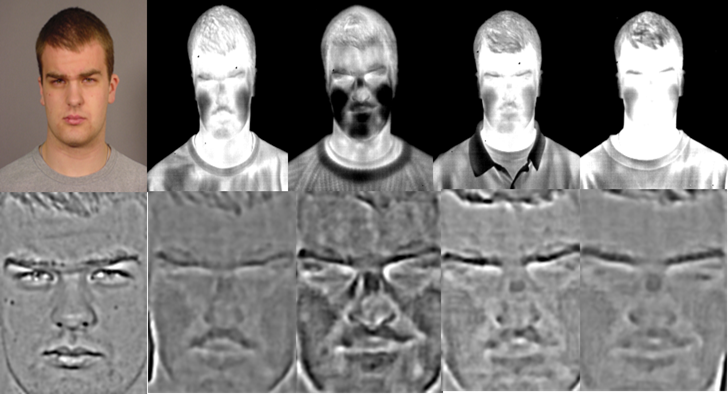}
    \caption{Aligned and preprocessed images of the corresponding visible and thermal images of a subject.}
    \label{fig:2a}
\end{figure}
\subsection{Implementation Details}
Here we provide the parameters used and related details for implementation. To compare the results to previous methods, we use a similar preprocessing for the thermal and visible images. We use the median filtering to remove the dead pixels in thermal images, then zero-mean normalization and afterwards used Difference of Gaussian `DoG' filtering to enhance edges and to further reduce the effect of lighting, Figure \ref{fig:2a} depicts aligned and preprocessed images of a subject. Dense features are then pooled from overlapping blocks in the image. The recognition performance varies between 2-5\% based on the descriptor used and the block size/overlap. A large overlap will produce more redundant information causing the network to saturate early while the block size effects the amount of structural information the network can still preserve while trying to map only the perceptual/modality difference. Based on our extensive evaluations we provide here the parameters that work well in practice. We have used a block size of $20\times20$ with a stride of $8$ pixels at two-scales (Gaussian smoothed with $\sigma=0.6$ and $\sigma=1$) of the image. As many prior methods \eg \cite{Klare13, hu2014, hu2015, choi2012} overwhelmingly stated the better performance of both the SIFT and HOG features in the thermal-visible face recognition. We experimented with both SIFT and HOG feature descriptors. However, here we present results with SIFT features as HOG produces $2-3\%$ inferior results than SIFT.

For the DPM model training, the number of units in the input/output and the hidden layers is important and varies the result. We use PCA to decorrelate and reduce the features to 64 dimensions. Each descriptor is further embedded with the block center position $(x,y)$, measured from the image center, to have the spatial position intact, this helps the model to better learn the local region specific transformations in the hidden layers. The number of units in the input/output layers are, therefore, $66$.

For the DPM network, we obtained better results with a minimum of 3-layers ($N=3$ in Equations \ref{eq1} and \ref{eq2}). Although, for comparison, we also report performance with using $N=2$ \ie one hidden layer. As our results show, the DPM mapping is highly effective even with this shallow DPM configuration. The minimum number of units in the hidden layers,  to ensure good results, is set empirically by using different combinations. Here, we report results using 200 units in each of the two hidden layers for the deep configuration. While in the case of shallow (1 hidden layer configuration), $1000$ hidden units in the single layer provide close results. It is worth noticing, that the the results vary slightly about 1-4\% using different number of hidden units in the range $200-1000$ in a layer. Finally the DPM is trained by pooling over 1 million patch vectors from the training set for each of the three datasets separately.
\subsection{Results}
We provide results of our evaluations using the settings described before. We first establish two baselines that would directly enable us to compare and appreciate the power of the proposed model. Along with these baselines we also compare the results of the proposed model with the previously published state of the art methods.
  
\noindent\textbf{Baseline-1:} As baseline-1 we use the same concatenated SIFT features without the DPM mapping. This would allow us to measure how much the learned mapping is contributing.

\noindent\textbf{Baseline-2:} As baseline-2 we use a state-of-the-art manifold learning method to embed the same SIFT vectors in a latent space using Partial Least Square (PLS) analysis. Vectors from both the thermal and visible images are projected through the learned PLS model in the latent space and then matched. The idea of PLS regression is to maximize the covariance between the dependant variable $Y$ and a weighted sum of the independent variable $X$ by finding a weighing vector $w$. In our application, $X$ and $Y$ corresponds to the the visible and thermal vectors respectively. To obtain the weight matrix the PLS model is built by decomposing the $(n\times m)$ input matrix $X\subset \mathbb{R}^m$ and the response matrix $Y\subset \mathbb{R}^m$ into
\begin{equation}
\begin{aligned}
X=TP^{T}+E \\
Y=UQ^{T}+F
\end{aligned}
\end{equation}
The $(n\times p)$ matrices $T$ and $U$ are called scores that contain the latent vectors, the $(m\times p)$ matrices $P$ and $Q$ are called loadings and $(n\times m)$ matrices $E$ and $F$ are the residuals. By using a greedy algorithm, we can iteratively obtain a set of weight vectors stored in a matrix $W$. At the end of each iteration, the matrices $X$ and $Y$ are deflated by subtracting their rank-one approximations based on $t$ and $u$ until the desired number of latent vectors $p$ is obtained. More details can be seen in \cite{rosipal2006}.

After obtaining $W$, we can compute the latent regression vectors from $X$ to $Y$ or vice versa by $Y=XB_{v}$ and $X=YB_{t}$ where the regression model $B$ is $B_{v}=W(P^{T}W)^{-1}$ and $B_{t}=W(Q^{T}W)^{-1}$ for the corresponding visible and thermal projections.

To ensure a proper baseline we perform PLS directly on the $m=66$ dimensional SIFT vectors space. The number of latent vectors is found by cross validation on the training set and is set to $p=20$ in all of our experiments. Each of the $66$ dimensional SIFT vector from visible and thermal image is projected to this $p=20$ dimensional latent space using the procedure mentioned above. All the projected vectors from an image are then concatenated and matched directly, similar to the procedure we adopt in baseline-1 and with the proposed DPM mapping.

Note that the previously published state of the art methods e.g., \cite{hu2015} also uses a similar PLS framework. While we also compare with their method, the difference in our PLS-baseline is apparent not only in the choice of feature vectors but also the way PLS model is learned. \cite{hu2015} use a discriminant PLS framework by building specifically discriminant PLS models for each identity in the database. 

With these baselines one can appreciate the benefit of using the proposed DPM mapping by directly comparing the results on the same features using 1) direct matching and 2) using a state of the art regression mapping in a latent space obtained by PLS by exploiting the same amount of training data.  

\subsubsection{Identification Evaluation}
\textbf{Results on UND-X1:} Table \ref{table:1} presents the comparison results (Rank-1 identification score) with the baseline and the best published state-of-the-art results on the UND-X1 dataset. We present results using three different gallery settings. The most difficult and restricted setting with only one neutral visible image per subject in the gallery, using two visible images per subject in the gallery (the first two images, neutral and with smiling expression) and using all the available visible images per subject as gallery. Note that all the available thermal images/subject are used as probes. The number of images for each subject in the thermal probe set varies from a minimum of $4$ to a maximum of $40$ depicting appearance variations due to expressions, lighting and time-lapse. As the results show, we improve the state-of-the-art best published results of \cite{hu2015} by 9\% in the 1-visible per subject gallery setting and more than 10\% in 2 per subject and all-available per subject cases. We also report the results using shallow DPM configuration (with 1-layer). Our results show that even this shallow DPM configuration well surpasses the best published results on this dataset.

\begin{figure}
  \begin{minipage}[t]{0.48\textwidth}  
    \includegraphics[width=\textwidth]{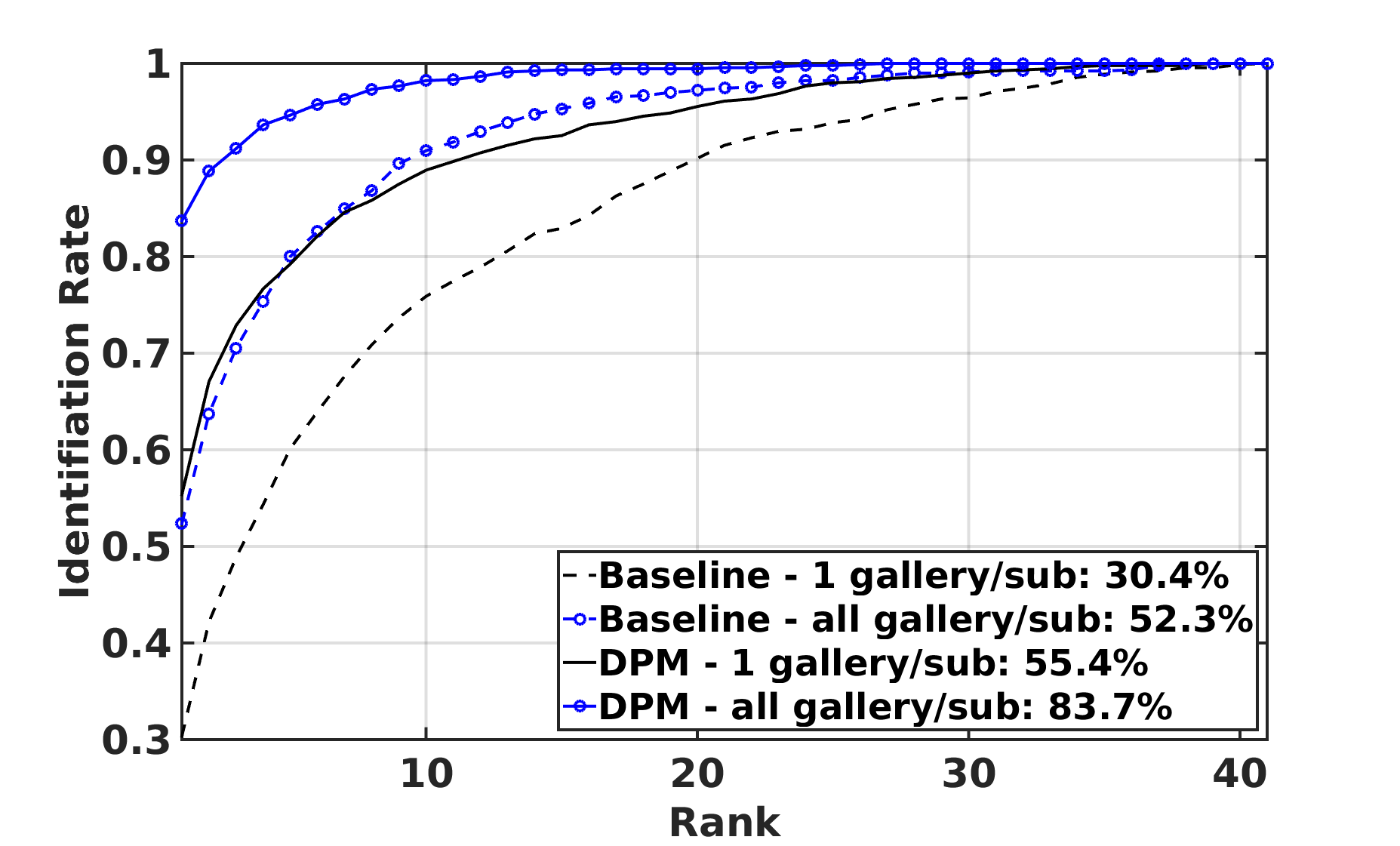}
	\caption{Rank-wise score on UND X1: comparison of baseline and DPM performance with 1-visible image/sub and all available visible images/sub in the gallery.}
    \label{fig:3}
  \end{minipage}
  \hfill
  \begin{minipage}[t]{0.48\textwidth}
    \includegraphics[width=\textwidth]{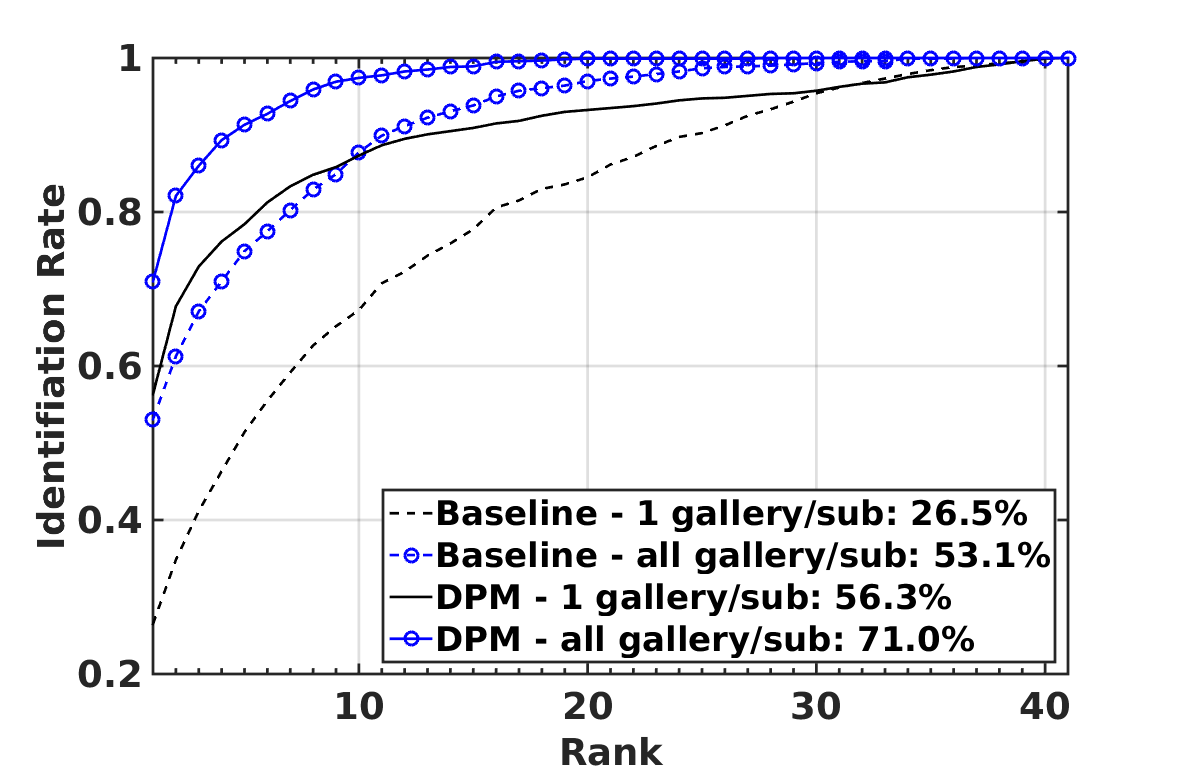}
\caption{Rank-wise score on Carl Dataset: comparison of baseline and DPM performance with 1-visible image/sub and all available visible images/sub in the gallery.}    
    %\caption{Verification performance on UND X1: all thermal probes with 2 visible image/sub in gallery. ROC curves from 1792 genuine and 71680 imposter attempts.}
    \label{fig:4}
  \end{minipage}
\end{figure}

Figure \ref{fig:3} presents the cumulative match characteristics `CMC' curves to measure the rank-wise identification performance of the presented method. It shows the effectiveness of the proposed DPM mapping in comparison with the baseline-1 (same features without the DPM mapping).
\begin{table}[t]
\centering
\begin{tabular}{@{}lccc@{}}
\toprule
                                 & \multicolumn{3}{c}{Gallery size: \# of visible images/subject} \\ \cmidrule(l){2-4} 
                                 & 1/subject         & 2/subject        & all available/subject   \\ \midrule
Baseline-1 (same features w/o DPM) & 30.36             & 35.60            & 52.34                   \\
Baseline-2 (PLS on same features) & 44.75             & 50.89            & 69.86                   \\
\cite{choi2012} & -                 & 49.9             & -                       \\
\cite{hu2015} & 46.3              & 50.4             & 72.7                    \\
DPM-\textit{1 hidden-layer} (ours)             & 50.67              & 58.26             & 79.57                   
\\
\textbf{DPM  (Ours)}             & \textbf{55.36}    & \textbf{60.83}   & \textbf{83.73}          \\ \bottomrule
\end{tabular}
\caption{Comparison of Rank-1 Identification accuracy (\%) on UND X1: using all thermal images as probes and visible images in the gallery.}
\label{table:1}
\end{table}

\begin{table}[t]
\centering
\begin{tabular}{@{}lccc@{}}
\toprule
                                 & \multicolumn{3}{c}{Gallery size: \# of visible images/subject} \\ \cmidrule(l){2-4} 
                                 & 1/subject         & 2/subject        & all available/subject   \\ \midrule
Baseline-1 (same features w/o DPM) & 26.50             & 33.83            & 53.08                   \\
Baseline-2 (PLS on same features) & 31.75             & 34.66            & 51.58                   \\
DPM-\textit{1 hidden-layer} (ours)             & 50.42              & 59.92             & 69.33                   
\\
\textbf{DPM  (ours)}             & \textbf{56.33}    & \textbf{60.08}   & \textbf{71.00}          \\ \bottomrule
\end{tabular}
\caption{Rank-1 Identification accuracy (\%) on Carl database: using all thermal images as probes and visible images in the gallery.}
\label{table:2}
\end{table}

\noindent\textbf{Results on Carl Database:} The Carl dataset is the most challenging dataset because of the present strong facial appearance variations due to different illumination changes , time lapse, expressions, alignment errors and very low resolution LWIR images ($160 \times 120$ pixels). To benchmark the full scale dataset, we present the identification results in a similar varying visible images/subject gallery setting. As mentioned earlier, the probe set contains all the thermal (LWIR) images from 21  test subjects and matched against a gallery containing visible images of all the 41 subjects. Table \ref{table:2} presents the obtained Rank-1 scores of both the baseline and the presented DPM mapping across different gallery image settings. Note while the Rank-1 scores on Carl dataset, despite a more difficult imaging scenario than UND-X1, are comparable in the 1 image per subject and 2 image per subject gallery setting. The relatively low Rank-1 score when using all available visible images per subject in the gallery is understandable considering the large number of gallery images (60 visible images/subject =2460 images form 41 subjects), where these visible images contain very difficult illumination variations (natural light, artificial light and infra-red illumination). We also tested the effect of overall gallery size i.e., only having visible images of the 21 test subjects in the gallery. Here, the Rank-1 scores using DPM improves as expected. We obtained an accuracy of 66.58\% with the 1-image per subject, 72.92\% with 2-images per subject and 82.50\% with all available images per subject gallery setting.

Figure \ref{fig:4} presents CMC curves to measure the rank-wise identification performance of the presented method and the baseline-1 on Carl dataset.
   
\noindent\textbf{Results on NVESD Dataset:} The NVESD dataset helps evaluate the method on both MWIR and LWIR thermal domains. Given the image acquisitions from relatively better thermal sensors ($640\times480$ pixel thermal images), the evaluations on NVESD resulted in much better performance. We evaluate and compare the Rank-1 identification results with that of the state-of-the-art results of \cite{hu2015}, using the same test protocol. 

We primarily evaluate the effect of exercise on the thermal signatures of the subjects across different subject-camera ranges (1m, 2m, and 4m). For this, the 25 subject subset both from MWIR and LWIR thermal sets is used as probe and matched against 2-visible images per subject in the gallery. The two visible-image per subject gallery images are the neutral mugshots, acquired at 1m range in before exercise condition. 
Following \cite{hu2015}, we prepare 6 probe sets for each MWIR and LWIR thermal test sets (before exercise and after exercise condition $\times$ three subject-to-camera ranges). Each probe set contains 50 images (2/subject) in each of the specific condition-range pair test. Table \ref{table:3} presents the Rank-1 scores for MWIR-Visible and LWIR-Visible scenario of this experiment. As the results depict, DPM mapping improves the state-of-the-art by a considerable margin, especially at longer ranges. An intriguing result is the better accuracy at 1m and 2m range of after exercise condition than before exercise. A closer look at the misclassified image/s of before-exercise probe sets at these ranges reveals that the effect seems to be with facial expressions in these pairs causing the similarity scores falling slightly below the rank-1 threshold. Given that the training has not seen the images in after-exercise condition, this also implies that the DPM mapping is not effected by the small thermal change present in the images due to this.

\begin{table}[t]
\begin{minipage}[t]{0.40\textwidth} 
\begin{tabular}{@{}lccc@{}}
\toprule
                                 & \multicolumn{3}{c}{Subject-to-camera range:} \\ \cmidrule(l){2-4} 
                                 & 1 m         & 2 m        & 4 m   \\ \midrule
Before Exercise 	&(92)\textbf{98}			&(70)\textbf{86}				&(58)\textbf{82} \\ 	 
\\
After Exercise             & (88)\textbf{100}              & (68)\textbf{92}             &(48)\textbf{74}                   
\\ \bottomrule
\end{tabular}
\begin{center} (a) \textbf{MWIR-Visible} \end{center}
%\caption{(a) MWIR-Visible}
\end{minipage}
%\end{table}
\hfill
%\begin{table}[t]
\begin{minipage}[t]{0.40\textwidth}
\begin{tabular}{@{}lccc@{}}
\toprule
                                 & \multicolumn{3}{c}{Subject-to-camera range:} \\ \cmidrule(l){2-4} 
                                 & 1 m         & 2 m        & 4 m   \\ \midrule
Before Exercise 	&(70)\textbf{94}			&(64)\textbf{92}				&(30)\textbf{84} \\ 	 
\\
After Exercise             & (64)\textbf{96}              & (62)\textbf{98}             &(28)\textbf{64}                   
\\ \bottomrule
\end{tabular}
\begin{center} (b) \textbf{LWIR-Visible} \end{center}
%\caption{(b) LWIR-Visible}
\end{minipage}
\caption{Rank-1 Identification accuracy (\%) on NVESD: Using Before Exercise and After Exercise thermal probe images at each of the three distance ranges. (a) Using MWIR probe images, (b) using LWIR probe images. Gallery includes 2 visible image/subject acquired in before exercise condition at 1 m range of all the 50 subjects. Parenthesis ($\cdot$) depict accuracies of the current best published results of \cite{hu2015}, Our results are depicted in bold.}
\label{table:3}
\end{table}

For MWIR-visible scenario we obtained on average a Rank-1 score of 88.6\% versus an average of 70.6\% of \cite{hu2015}. We, therefore, improve the previous results by 18\% on the NVESD's MWIR-visible dataset. Similarly for LWIR-visible, we obtained on average Rank-1 score of 88\% versus an average of 53\% of \cite{hu2015}. Thus, on the NVESD's LWIR-visible dataset, we improve the result by 35\%. These results show that the learned DPM mapping is relatively stable and effective against the thermal signature change due to exercise while performs significantly better on image acquisition at longer range. Table \ref{table:6} and Table \ref{table:5} provide the overall Rank-1 accuracy of the Baselines and the DPM in different visible-image/subject in the gallery settings on MWIR-visible and LWIR-visbile NVESD datasets.

\begin{figure}
  \begin{minipage}[t]{0.48\textwidth}  
    \includegraphics[width=\textwidth]{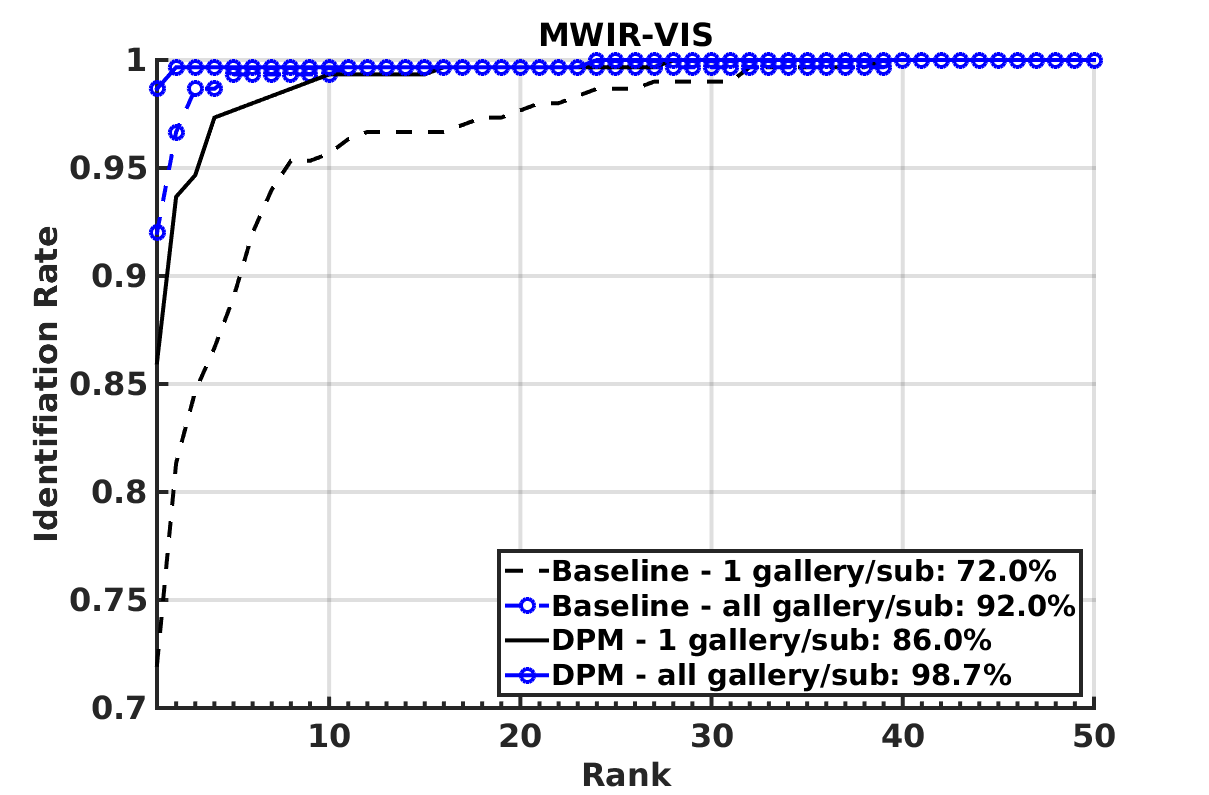}
	\caption{Rank-wise score on NVESD MWIR-visible dataset: comparison of baseline and DPM performance with 1-visible image/sub and all available visible images/sub in the gallery.}
    \label{fig:5}
  \end{minipage}
  \hfill
  \begin{minipage}[t]{0.48\textwidth}
    \includegraphics[width=\textwidth]{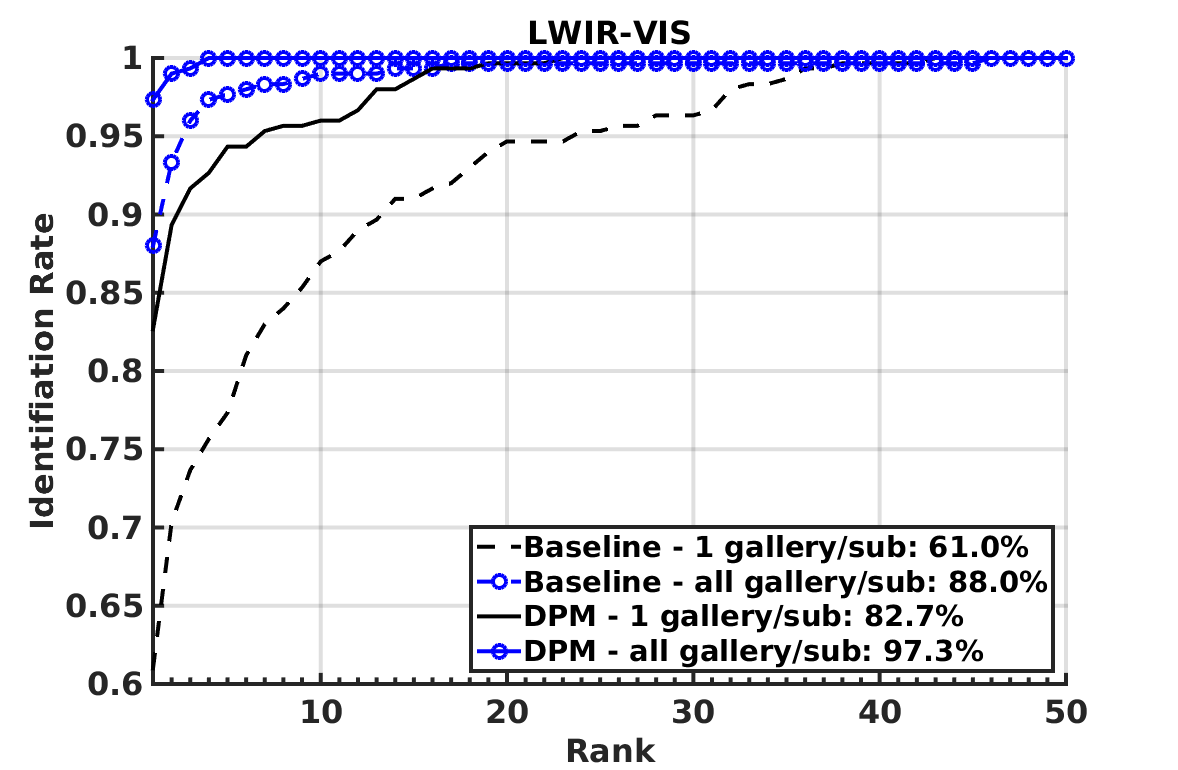}
\caption{Rank-wise score on NVESD LWIR-visible dataset: comparison of baseline and DPM performance with 1-visible image/sub and all available visible images/sub in the gallery.}    
  \label{fig:6}
  \end{minipage}
\end{figure}

Figure \ref{fig:5} and Figure \ref{fig:6} provide the similar rank-wise performance of our method and compares it with the baseline-1 features in the 1-visible per subject and the all-available visible per subject gallery setting.
\begin{table}[t]
\centering
\begin{tabular}{@{}lccc@{}}
\toprule
                                 & \multicolumn{3}{c}{Gallery size: \# of visible images/subject} \\ \cmidrule(l){2-4} 
                                 & 1/subject         & 2/subject        & all available/subject   \\ \midrule
Baseline-1 (same features w/o DPM) & 72.00             & 75.66            & 92.00                   \\
Baseline-2 (PLS on same features) & 77.66             & 79.33            & 97.66                  \\
%\cite{hu2015} & -              & 70.6             & -                    \\
DPM-\textit{1 hidden-layer} (ours)             & \textbf{86.66}              & 87.33             & 97.66                   
\\
\textbf{DPM  (Ours)}             & 86.00    & \textbf{87.66}   & \textbf{98.66}          \\ \bottomrule
\end{tabular}
\caption{\textbf{MWIR-Visible} - Overall Rank-1 Identification accuracy (\%) on NVESD: using all MWIR images as probes and visible images in the gallery.}
\label{table:6}
\end{table}

\begin{table}[t]
\centering
\begin{tabular}{@{}lccc@{}}
\toprule
                                 & \multicolumn{3}{c}{Gallery size: \# of visible images/subject} \\ \cmidrule(l){2-4} 
                                 & 1/subject         & 2/subject        & all available/subject   \\ \midrule
Baseline-1 (same features w/o DPM) & 61.00             & 64.33            & 88.00                   \\
Baseline-2 (PLS on same features) & 66.66             & 67.00            & 93.66                  \\
%\cite{hu2015} & -              & 53.0             & -                    \\
DPM-\textit{1 hidden-layer} (ours)             & \textbf{83.67}              & 82.00             & 96.67                   
\\
\textbf{DPM  (Ours)}             & 82.67    & \textbf{85.00}   & \textbf{97.33}          \\ \bottomrule
\end{tabular}
\caption{\textbf{LWIR-Visible} - Overall Rank-1 Identification accuracy (\%) on NVESD: using all LWIR images as probes and visible images in the gallery.}
\label{table:5}
\end{table}

\subsubsection{Verification Evaluation}
\begin{figure}
  \begin{minipage}[t]{0.48\textwidth}  
    \includegraphics[width=\textwidth]{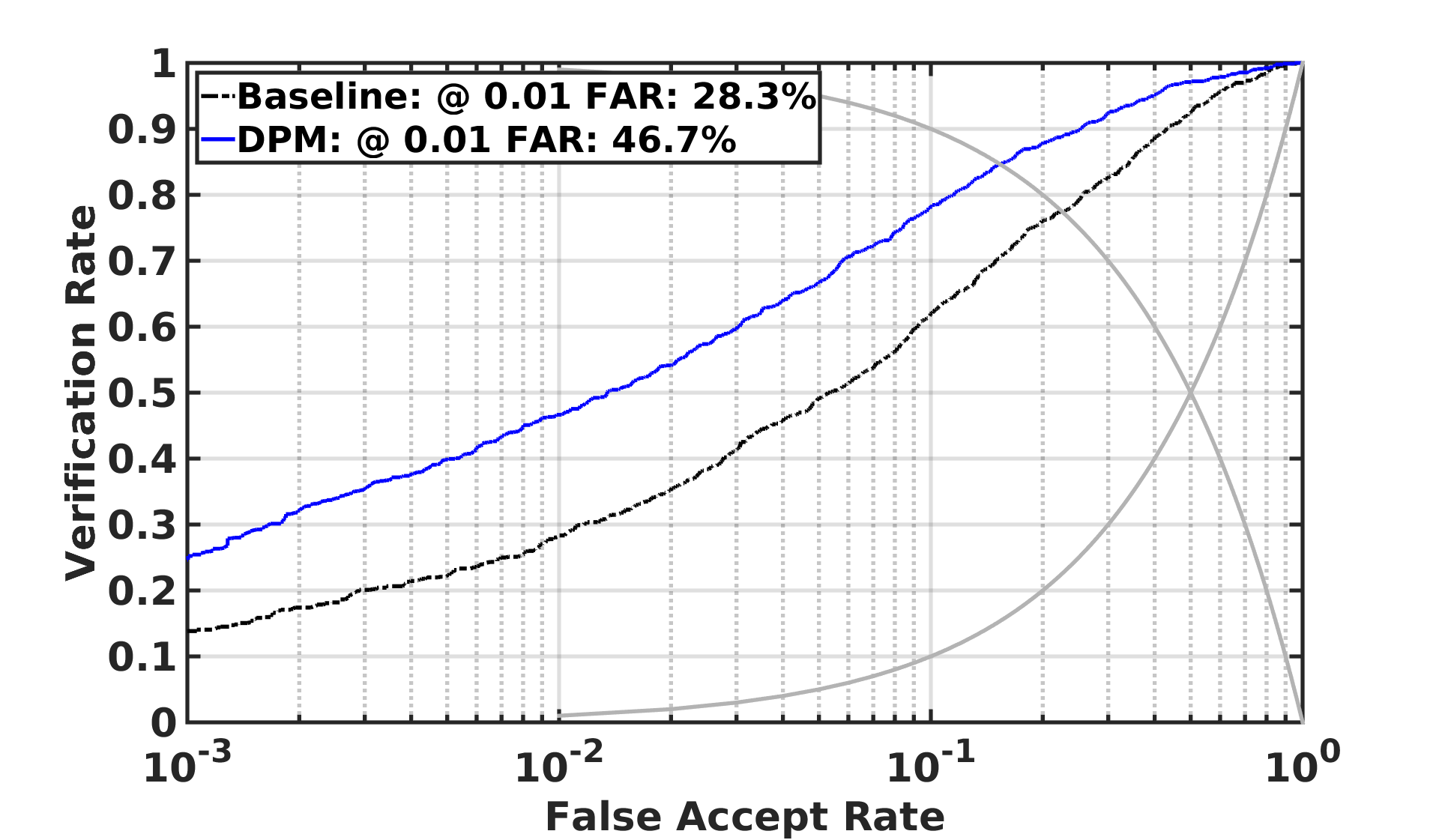}
	\caption{Verification performance on UND X1: all thermal probes with 2 visible image/sub in gallery.}
    \label{fig:7}
  \end{minipage}
  \hfill
  \begin{minipage}[t]{0.48\textwidth}
    \includegraphics[width=\textwidth]{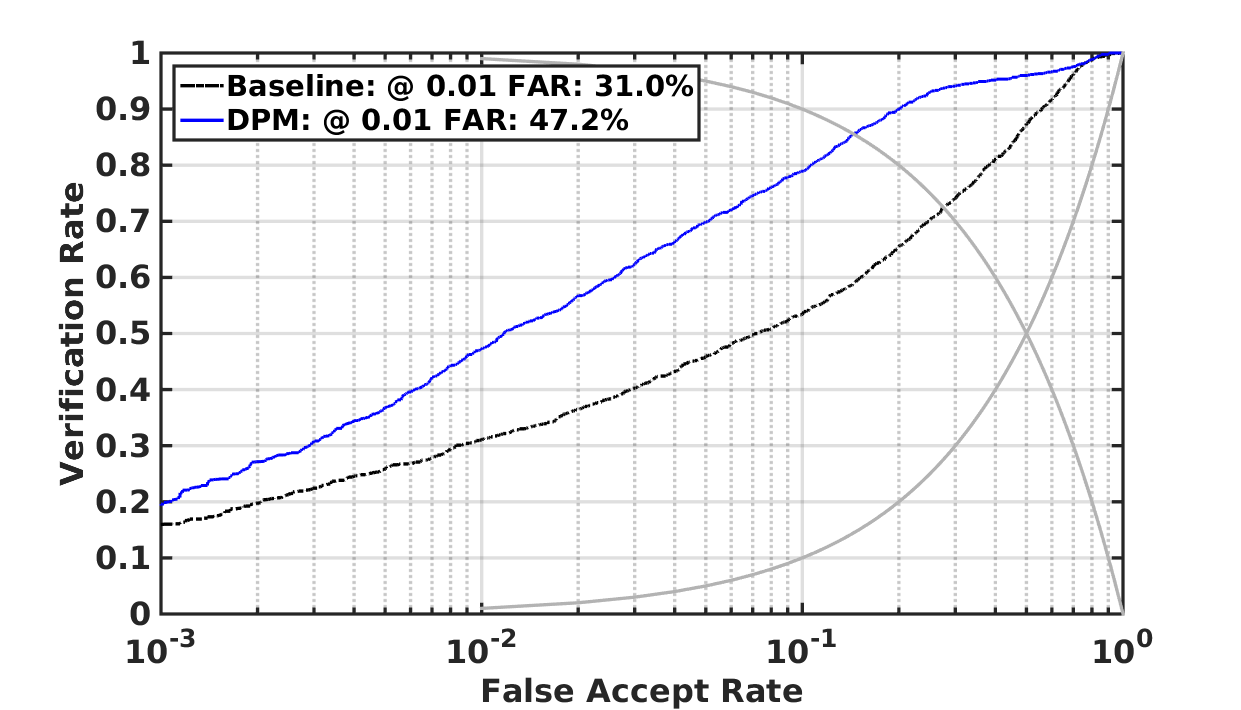}
\caption{Verification performance on Carl Database: all thermal probes with 2 visible image/sub in gallery.}    
        \label{fig:8}
  \end{minipage}
\end{figure}

\begin{figure}
  \begin{minipage}[t]{0.48\textwidth}  
    \includegraphics[width=\textwidth]{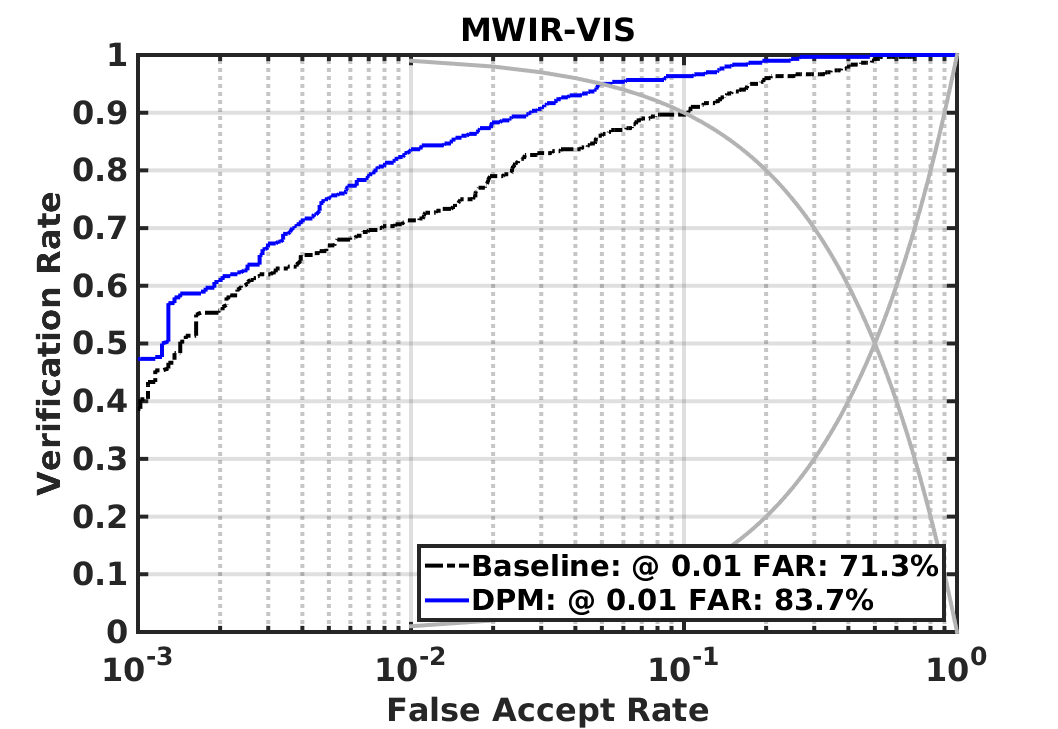}
	\caption{Verification performance on NVESD MWIR dataset: all thermal probes with 2 visible image/sub in gallery.}
    \label{fig:9}
  \end{minipage}
  \hfill
  \begin{minipage}[t]{0.48\textwidth}
    \includegraphics[width=\textwidth]{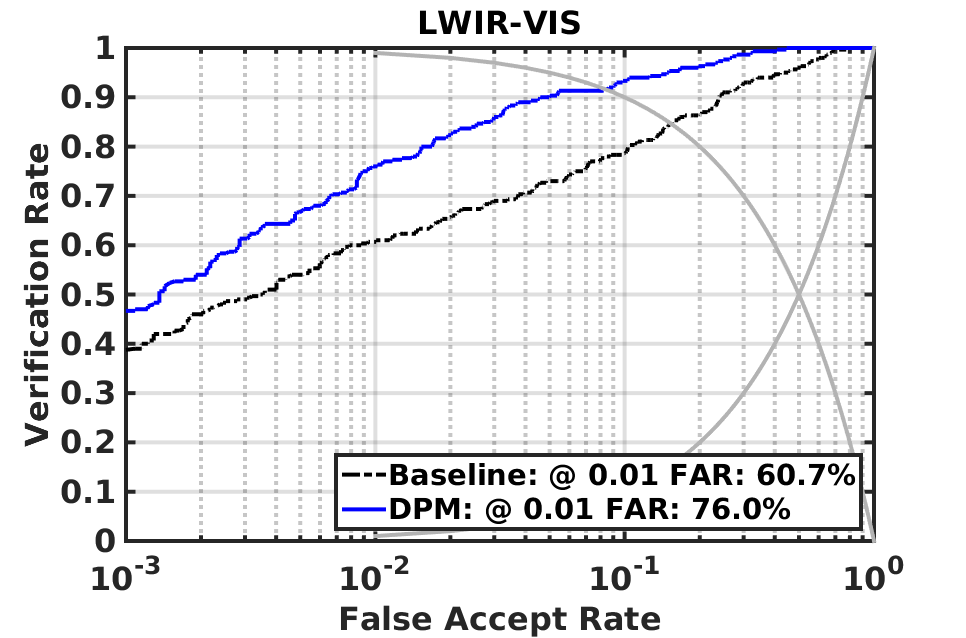}
\caption{Verification performance on NVESD LWIR dataset: all thermal probes with 2 visible image/sub in gallery.}    
        \label{fig:10}
  \end{minipage}
\end{figure}

We also evaluate the verification accuracy of such cross-modal scenario. We report here the results by using 2 visible images/subject in the gallery on UND-X1, Carl and NVESD datasets. Given the size of our thermal probe set, this amounts to having $1792$ genuine and $71680$ imposter attempts on UND-X1, $2520$ genuine and $100800$ imposter attempts on Carl database and $600$ genuine and $29400$ imposter attempts on both NVESD's LWIR and MWIR datasets. Figures \ref{fig:7}, \ref{fig:8}, \ref{fig:9} and \ref{fig:10} present the ROC curves measuring the verification performance gain over the baseline-1 using the DPM mapping.

\subsubsection{Effect of modality gap}
Finally, we present the experiment to measure the effect of the modality gap. Keeping everything fixed \ie, using the same baseline features and settings, we compute the Rank-1 identification score within the same modality. We use one-thermal image per subject (the neutral frontal image) to form the thermal gallery and test against all the remaining thermal images as probe. This is the same setting as we have used in the cross-modal thermal-visible case. Table \ref{table:modgap} compares the results of Thermal-Thermal and Thermal-Visible identification on all three datasets to quantify the effect of modality gap.


On UND-X1, we obtain a Rank-1 score of $89.7\%$ in the Thermal-Thermal identification scenario. While the rank-1 identification in the corresponding Thermal-Visible scenario (using the same baseline features) is $30.3\%$. This amounts to performance drop, purely due to modality change, of about $59\%$. This reflects the challenging nature of the problem and the existing research gap to tackle this. With DPM on the same features, the performance is improved by $25\%$. This amounts to bridging the existing modality gap of $59\%$ by more than $40\%$ on UND-X1. Similar observations can be made by interpreting the results on other datasets as given in Table \ref{table:modgap}.
\subsubsection{Computational Time}
Training the DPM on 12 cores 3.2-GHz CPU takes between $1-1.5$ hours on MATLAB. Preprocessing, features extraction and mapping using DPM only takes $45 ms$ for one image. This is even less in the testing case since no mapping is required for thermal images. At test time identifying one probe only takes $35 ms$. Since we are using just the dot product between the extracted probe vector and the gallery set, this is therefore very fast and capable of running in real-time at $\sim 28$ fps.

\begin{table}[t]
\centering
\begin{tabular}{@{}ccccc@{}}
\toprule
\multicolumn{5}{r}{Effect of Modality gap: Performance with 1 Gallery image/subject} \\ \cmidrule(l){2-5}
&Therm-Therm   & Therm-Vis  & Therm-Vis(DPM)  & Modal-gap bridged  \\ 
\midrule
 UND-X1 		&89.4             & 30.3            & 55.3                       & $\sim 42 \%$               \\
 Carl Dataset 	&61.7             & 26.5            & 56.3                       & $\sim 84 \%$               \\
NVESD MWIR 	&98.6             & 72            & 86                       & $\sim 48 \%$               \\
NVESD LWIR 	&97             & 61            & 82.6                       & $\sim 40 \%$                
%\bottomrule
\end{tabular}
\caption{Performance drop due to Modality gap: Rank-1 identification using 1 image/subject as gallery in Thermal-Thermal and Thermal-Visible matching using baseline features.}

\label{table:modgap}
\end{table}

\section{Discussion \& Conclusions}
\label{Disc}
While providing quantitative evidence of the effectiveness of the proposed approach, we can also compare our results qualitatively with some other methods that evaluate the thermal-visible face recognition on similar private datasets. \citep{bourlai2012} reported results on a MWIR-visible face recognition task using a $39$ subject dataset with four visible images per subject and three MWIR images per subject as probes with $1024\times1024$ pixel resolution. They reported a rank-1 performance of $53.9\%$. Our result of 86\% on the NVESD MWIR-visible dataset with just one visible image per subject in the gallery can be qualitatively compared to these results. Similarly, \cite{Chen2015} evaluated their method on a subset of the Carl dataset with 5 visible image per subject in the gallery and 5 thermal image per subject as probe. They, however, evaluate in a cross dataset scenario \ie, subspaces learned on another similar private dataset and tested on the Carl data subset. They reported a best rank-1 performance of 75.6\%. Compared to this, we evaluate the full scale Carl dataset with 60 thermal images per subject in the probe set. Our results with just 2 visible image per subject and multiple visible images per subject in the gallery, as depicted in Table \ref{table:2}, provide a qualitative motivation for the effectiveness of the proposed method. Finally, in a very recent proposal (\cite{Riggen2016}) the authors evaluate their own implementation of our presented DPM method on polarimetric thermal to visible face recognition problem achieving one of the highest performance on this dataset. This further substantiates the applicability of the presented method on a general cross modal image matching scenario. 

Given the very high recognition ability of Convolutional Neural Network (CNN) features in the conventional visible face recognition, a possible future direction to further improve these results is to directly train a CNN model. On the cross-modal face matching problem,  the CNN model can be trained in a similar metric learning fashion as presented by Google's FaceNet (\cite{schroff2015facenet}) or Oxford's VGG deep face model (\cite{Parkhi15}). The current limitation for a meaningful training of such a model is the lack of enough thermal facial images. It is hoped that, provided, the vast and demanding application of cross-modal face matching and availability of low-priced accessible thermal sensors, more and more thermal facial data can be acquired. Note, a direct application of the current pre-trained CNN models on the thermal images do not yield meaningful results. These models are trained on three channel RGB colour images where as the thermal images are single channel 16-bit images. Our initial investigations to get a baseline using the pre-trained CNN models of (\cite{schroff2015facenet} and \cite{Parkhi15}), by just replicating the thermal image in three channels, did not provide acceptable results. Here, however, a similar perceptual mapping strategy may be further investigated by training a CNN using only the image patches from both domains. Such  a direction may provide interesting results even with relatively less training data.
   
Conclusively, thermal-visible face recognition is a very difficult problem due to the inherent large modality difference. Our presented method is very effective and has benefits for many related applied computer vision and domain adaptation problems from image matching, detection to recognition. Similarly it is also very attractive for remote sensing applications, where it is a common problem to register and match images from different modalities (\eg, coming from different satellites). This can be used as a very effective prior step to bridge the modality and/or large resolution difference before using the representation vectors in any common learning scheme. The presented DPM approach is very useful, easy to train, and real-time capable due to little computational overhead and it provides a practical solution for a large related application industry. 

%\begin{acknowledgements}
%\end{acknowledgements}

\bibliographystyle{spbasic}      
\bibliography{bibours}  

\begin{thebibliography}{29}
\providecommand{\natexlab}[1]{#1}
\providecommand{\url}[1]{{#1}}
\providecommand{\urlprefix}{URL }
\expandafter\ifx\csname urlstyle\endcsname\relax
  \providecommand{\doi}[1]{DOI~\discretionary{}{}{}#1}\else
  \providecommand{\doi}{DOI~\discretionary{}{}{}\begingroup
  \urlstyle{rm}\Url}\fi
\providecommand{\eprint}[2][]{\url{#2}}

\bibitem[{Bourlai et~al(2012)Bourlai, Ross, Chen, and Hornak}]{bourlai2012}
Bourlai T, Ross A, Chen C, Hornak L (2012) A study on using mid-wave infrared
  images for face recognition. In: SPIE Defense, Security, and Sensing,
  International Society for Optics and Photonics, pp 83,711K--83,711K

\bibitem[{Byrd(2013)}]{NVESD}
Byrd K (2013) Preview of the newly acquired nvesd-arl multimodal face database.
  In: Proc. SPIE, vol 8734, pp 8734--34

\bibitem[{Chen and Ross(2015)}]{Chen2015}
Chen C, Ross A (2015) Matching thermal to visible face images using hidden
  factor analysis in a cascaded subspace learning framework. Pattern
  Recognition Letters \doi{http://dx.doi.org/10.1016/j.patrec.2015.06.021},
  \urlprefix\url{http://www.sciencedirect.com/science/article/pii/S0167865515001932}

\bibitem[{Chen et~al(2005)Chen, Flynn, and Bowyer}]{chen2005ir}
Chen X, Flynn PJ, Bowyer KW (2005) Ir and visible light face recognition.
  Computer Vision and Image Understanding 99(3):332--358

\bibitem[{Choi et~al(2012)Choi, Hu, Young, and Davis}]{choi2012}
Choi J, Hu S, Young SS, Davis LS (2012) Thermal to visible face recognition.
  In: SPIE Defense, Security, and Sensing, International Society for Optics and
  Photonics, pp 83,711L--83,711L

\bibitem[{Espinosa-Duró et~al(2013)Espinosa-Duró, Faundez-Zanuy, and
  Mekyska}]{carlDataset}
Espinosa-Duró V, Faundez-Zanuy M, Mekyska J (2013) A new face database
  simultaneously acquired in visible, near-infrared and thermal spectrums.
  Cognitive Computation 5(1):119--135, \doi{10.1007/s12559-012-9163-2},
  \urlprefix\url{http://dx.doi.org/10.1007/s12559-012-9163-2}

\bibitem[{Ganin and Lempitsky(2014)}]{ganin2014}
Ganin Y, Lempitsky V (2014) Unsupervised domain adaptation by backpropagation.
  arXiv preprint arXiv:14097495

\bibitem[{Glorot and Bengio(2010)}]{glorot2010}
Glorot X, Bengio Y (2010) Understanding the difficulty of training deep
  feedforward neural networks. In: International conference on artificial
  intelligence and statistics, pp 249--256

\bibitem[{Hu et~al(2014{\natexlab{a}})Hu, Lu, and Tan}]{hu2014discriminative}
Hu J, Lu J, Tan YP (2014{\natexlab{a}}) Discriminative deep metric learning for
  face verification in the wild. In: Computer Vision and Pattern Recognition
  (CVPR), 2014 IEEE Conference on, IEEE, pp 1875--1882

\bibitem[{Hu et~al(2014{\natexlab{b}})Hu, Gurram, Kwon, and Chan}]{hu2014}
Hu S, Gurram P, Kwon H, Chan AL (2014{\natexlab{b}}) Thermal-to-visible face
  recognition using multiple kernel learning. In: SPIE Defense+ Security,
  International Society for Optics and Photonics, pp 909,110--909,110

\bibitem[{Hu et~al(2015)Hu, Choi, Chan, and Schwartz}]{hu2015}
Hu S, Choi J, Chan AL, Schwartz WR (2015) Thermal-to-visible face recognition
  using partial least squares. Journal of the Optical Society of America `JOSA'
  A 32(3):431--442

\bibitem[{Klare and Jain(2010)}]{Klare2010}
Klare B, Jain A (2010) Heterogeneous face recognition: Matching nir to visible
  light images. In: Pattern Recognition (ICPR), 2010 20th International
  Conference on, pp 1513--1516

\bibitem[{Klare and Jain(2013)}]{Klare13}
Klare BF, Jain AK (2013) Heterogeneous face recognition using kernel prototype
  similarities. Pattern Analysis and Machine Intelligence, IEEE Transactions on
  35(6):1410--1422

\bibitem[{Lei and Li(2009)}]{lei2009}
Lei Z, Li SZ (2009) Coupled spectral regression for matching heterogeneous
  faces. In: Computer Vision and Pattern Recognition, 2009. CVPR 2009. IEEE
  Conference on, IEEE, pp 1123--1128

\bibitem[{Li et~al(2008)Li, Hao, Zhang, and Dou}]{li2008}
Li J, Hao P, Zhang C, Dou M (2008) Hallucinating faces from thermal infrared
  images. In: Image Processing, 2008. ICIP 2008. 15th IEEE International
  Conference on, IEEE, pp 465--468

\bibitem[{Li et~al(2007)Li, Chu, Liao, and Zhang}]{li2007}
Li SZ, Chu R, Liao S, Zhang L (2007) Illumination invariant face recognition
  using near-infrared images. Pattern Analysis and Machine Intelligence, IEEE
  Transactions on 29(4):627--639

\bibitem[{Liao et~al(2009)Liao, Yi, Lei, Qin, and Li}]{liao2009}
Liao S, Yi D, Lei Z, Qin R, Li SZ (2009) Heterogeneous face recognition from
  local structures of normalized appearance. In: Advances in Biometrics,
  Springer, pp 209--218

\bibitem[{Mostafa et~al(2013)Mostafa, Hammoud, Ali, and Farag}]{mostafa2013}
Mostafa E, Hammoud R, Ali A, Farag A (2013) Face recognition in low resolution
  thermal images. Computer Vision and Image Understanding 117(12):1689--1694

\bibitem[{Nicolo and Schmid(2012)}]{nicolo2012}
Nicolo F, Schmid NA (2012) Long range cross-spectral face recognition: Matching
  swir against visible light images. Information Forensics and Security, IEEE
  Transactions on 7(6):1717--1726

\bibitem[{Parkhi et~al(2015)Parkhi, Vedaldi, and Zisserman}]{Parkhi15}
Parkhi OM, Vedaldi A, Zisserman A (2015) Deep face recognition. In: British
  Machine Vision Conference

\bibitem[{Riggan et~al(2016)Riggan, Nathaniel, and Shuowen}]{Riggen2016}
Riggan BS, Nathaniel JS, Shuowen H (2016) Optimal feature learning and
  discriminative framework for polarimetric thermal to visible face
  recognition. In: IEEE Winter Conference on Applications of Computer Vision
  (WACV)

\bibitem[{Rosipal and Kr{\"a}mer(2006)}]{rosipal2006}
Rosipal R, Kr{\"a}mer N (2006) Overview and recent advances in partial least
  squares. In: Subspace, latent structure and feature selection, Springer, pp
  34--51

\bibitem[{Ross and Hornak(2010)}]{ross2010}
Ross TBNKA, Hornak BCL (2010) Cross-spectral face verification in the short
  wave infrared (swir) band

\bibitem[{Sarfraz and Stiefelhagen(2015)}]{sarfraz2015}
Sarfraz MS, Stiefelhagen R (2015) Deep perceptual mapping for thermal to
  visible face recognition. In: British Machine Vision Conference

\bibitem[{Schroff et~al(2015)Schroff, Kalenichenko, and
  Philbin}]{schroff2015facenet}
Schroff F, Kalenichenko D, Philbin J (2015) Facenet: A unified embedding for
  face recognition and clustering. arXiv preprint arXiv:150303832

\bibitem[{Socolinsky and Selinger(2002)}]{socolinsky2002}
Socolinsky DA, Selinger A (2002) A comparative analysis of face recognition
  performance with visible and thermal infrared imagery. Tech. rep., DTIC
  Document

\bibitem[{Taigman et~al(2014)Taigman, Yang, Ranzato, and Wolf}]{taigman2014}
Taigman Y, Yang M, Ranzato M, Wolf L (2014) Deepface: Closing the gap to
  human-level performance in face verification. In: Computer Vision and Pattern
  Recognition (CVPR), 2014 IEEE Conference on, IEEE, pp 1701--1708

\bibitem[{Yi et~al(2007)Yi, Liu, Chu, Lei, and Li}]{yi2007}
Yi D, Liu R, Chu R, Lei Z, Li SZ (2007) Face matching between near infrared and
  visible light images. In: Advances in Biometrics, Springer, pp 523--530

\bibitem[{Zhou et~al(2014)Zhou, Mian, Wei, Creighton, Hossny, and
  Nahavandi}]{zhou2014recent}
Zhou H, Mian A, Wei L, Creighton D, Hossny M, Nahavandi S (2014) Recent
  advances on singlemodal and multimodal face recognition: A survey. Human
  Machine Systems, IEEE Transactions on 44(6):701--716

\end{thebibliography}

\end{document}